\documentclass[letterpaper, 10 pt, conference]{ieeeconf}  

\IEEEoverridecommandlockouts                              

\usepackage{graphics} 
\usepackage{epsfig} 
\usepackage{times} 
\usepackage{amsmath} 
\usepackage{amssymb}  
\usepackage{algorithm}
\usepackage{algorithmicx}
\usepackage{algpseudocode}
\usepackage{cite}

\usepackage{cleveref}

\title{\LARGE \bf
Online Imitation Learning for Manipulation via Decaying Relative Correction through Teleoperation}

\author{Cheng Pan$^{*1}$, Hung Hon Cheng$^{*1}$ \& Josie Hughes$^{1}$ 
\thanks{$^{1}$The authors are with Faculty of Mechanical Engineering, Swiss Federal Institute of Technology Lausanne, 1015 Lausanne, Switzerland {\tt\small cheng.pan@epfl.ch}}%
\thanks{Note: This work has been submitted to the IEEE for possible publication. Copyright may be transferred without notice, after which this version may no longer be accessible.}%
}

\begin{document}

\maketitle

\begin{abstract}

Teleoperated robotic manipulators enable the collection of demonstration data, which can be used to train control policies through imitation learning. However, such methods can require significant amounts of training data to develop robust policies or adapt them to new and unseen tasks. While expert feedback can significantly enhance policy performance, providing continuous feedback can be cognitively demanding and time-consuming for experts.
To address this challenge, we propose to use a cable-driven teleoperation system which can provide spatial corrections with 6 degree of freedom to the trajectories generated by a policy model.
Specifically, we propose a correction method termed Decaying Relative Correction (DRC) which is based upon the spatial offset vector provided by the expert and exists temporarily, and which reduces the intervention steps required by an expert.
Our results demonstrate that DRC reduces the required expert intervention rate by 30\% compared to a standard absolute corrective method. Furthermore, we show that integrating DRC within an online imitation learning framework rapidly increases the success rate of manipulation tasks such as raspberry harvesting and cloth wiping.

\end{abstract}


\section{Introduction}






\begin{figure}[tb]
    \centering    \includegraphics[width=0.97\columnwidth]{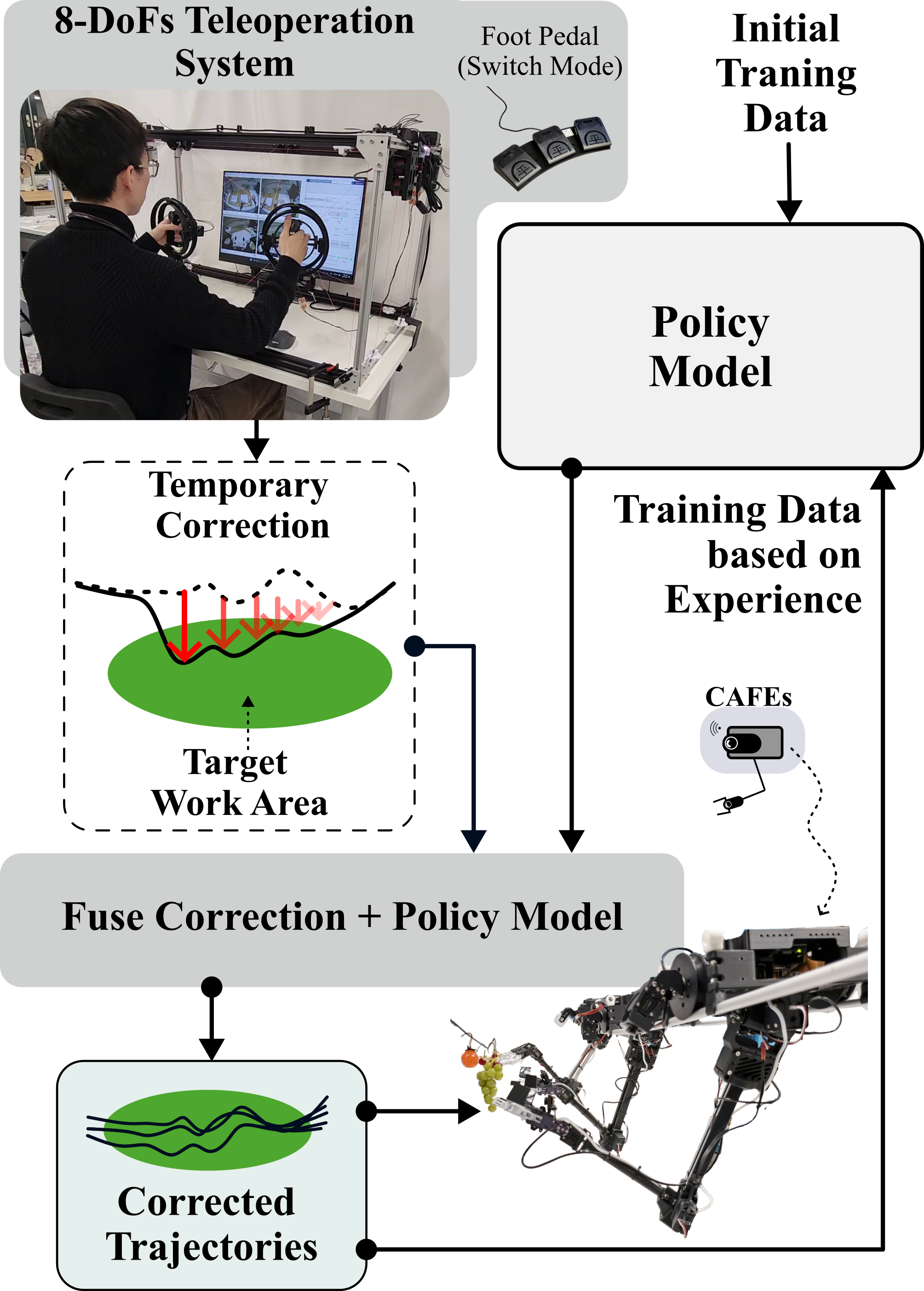}
    \caption{Architecture of our proposed approach. The initial training data for online imitation learning is first collected through expert demonstration. The trained policy model is then deployed to robot, but still allowing the operator to correct motions in real time. Based on the corrected trajectories, the model is retrained and redeployed iteratively }
    \label{fig:First_fig}
    \vspace{-1em}
\end{figure}

Robotic manipulation is a core challenge in robotics, requiring precise control and generalizability to diverse environments and tasks. 
Beyond developing hardware which has dexterous capabilities, we require methods for rapidly teaching or demonstrating  manipulation tasks to robots ~\cite{kroemer2021review}.
Imitation learning is a widely used approach for teaching robots and leverages expert demonstrations to train a control policy~\cite{fang2019survey}. However, the quantity of demonstrations is often limited or only cover a limited portion of the environment and action spaces that the robot may encounter during real-world operation~\cite{osa2018algorithmic}. As a result, the policy can fail to generalize, and the robot can make mistakes when encountering unseen or new scenarios which show variation to the initial demonstrations. Incorporating corrective feedback from experts could enables continuous improvement of the robot policy in an online manner, enhances its performance and robustness \cite{celemin2022interactive}.

Providing correction or feedback in online human-interactive imitation learning requires significant attention and effort from human experts \cite{hoque2021lazydagger}. Most existing methods rely on experts providing absolute corrections in the form of precise action or state values for each incorrect steps \cite{ross2011reduction, celemin2022interactive, kelly2019hg}, which can be cognitively demanding and exhausting for human operators. 
Alternatively, preference-based \cite{brown2019extrapolating} and evaluative feedback \cite{chisari2022correct} methods simplify the feedback process by requiring only a single feedback instance from the expert. However, these approaches often suffer from ambiguity in corrections, necessitating a large number of feedback iterations for effective policy learning \cite{brown2019extrapolating, christiano2017deep}.
In comparison, relative correction for trajectories offers a more natural and efficient way to refine trajectories by specifying adjustments such as shifting position, increasing speed, or modifying force \cite{celemin2015coach, celemin2022interactive, celemin2019interactive}. Works in \cite{celemin2015coach, perez2020interactive} utilize binary signals to increase/decrease the robot actions to be corrected. \cite{jauhri2020interactive} generalizes this method to the relative correction of the state space. However, in these works relative corrections was only provided by a binary signals, which is neither efficient nor intuitive for human operators. The lack of using of signals with continuous magnitude can be attributed to the absence of an intuitive hardware interface. Keyboard or other buttons devices are typically usually used to provide binary corrective signals in these previous works \cite{celemin2015coach, jauhri2020interactive, perez2020interactive}. Additionally, human feedback is typically applied only to the current state or action, and the corrective inputs are required from experts for all inappropriate state or action steps \cite{celemin2015coach, perez2020interactive, jauhri2020interactive}, limiting its efficiency.

We propose a Decaying Relative Correction (DRC) method for correction of a robotic manipulation, which can be fused with a policy model to require only minimal intervention from an expert. 
This correction mode is made possible by a use of 8 degree of freedom (DoF) teleoperation system and cable driven robot.
The relative correction motion provided by the experts teleoperation provides the vector of correction which is is only temporary, and stays valid for only a period of time, disappearing with a given decay rate. We hypothesize that DRC from human experts is more efficient in terms of the number of required intervention steps compared to standard absolute correction. Since trends in relative corrections (e.g., continuous adjustments over time) could remain valid for subsequent states or actions, leveraging this temporal consistency is crucial for improving the efficiency of online imitation learning systems.
In our robotic system, DRC from the expert is represented as a spatial difference vector, which is provided by the expert activating our cable-driven teleoperation handle (see Fig. 1). This cable handle enables human experts to provide corrective signals with continuous magnitudes, allowing intuitive corrective teleportation for online imitation learning.
The efficiency and intuitiveness of the proposed correction method is expected to enable single human expert to supervise multiple robots and provide corrections.

Our proposed DRC and cable-handle based online imitation learning framework is illustrated in Fig. \ref{fig:First_fig}. The robots are firstly deployed using a policy model which is pretrained using a standard imitation learning (behavior cloning). During the deployment, the human expert provide DRC signals to the robot  via the cable teleoperation handle when the robot make mistakes. The corrected trajectories will be saved to update the policy model of robots. 

We demonstrate the use of the DRC signals on our robot and benchmark against a direct 'absolute' override approach, demonstrating that this requires more input from an expert. 
Using the DRC we perform iterative online training based on DRC-collected new trajectories in an online imitation learning framework.  Using this show how the success rate of two tasks (rapsberry harvesting and wiping with a cloth) increases from 30\% to above 80\%. Finally, we use the same method to show how online correction can be used to adapt the policy model to unseen objects or tasks.





In the remainder of this paper we first introduce the methods, providing details of the robotic setup and the proposed DRC method and online imitation learning framework. The experimental setup and results are given, before finishing with a discussion of the contributions and future outlook.

\begin{figure}[!t]
     \centering
     \vspace{0.3cm}
     \includegraphics[width=0.9\columnwidth]{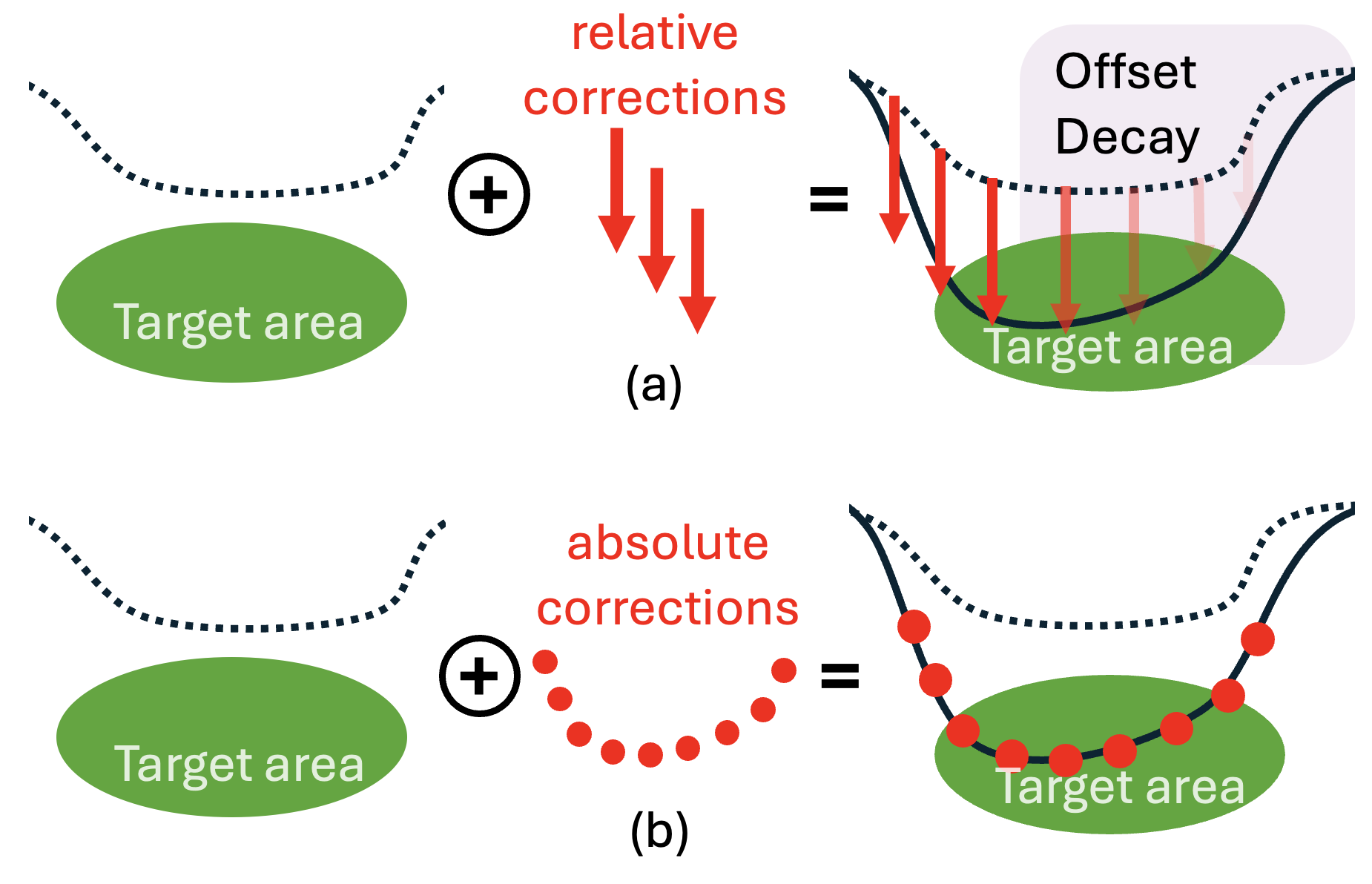}
     \caption{Two types of correction methods: (a) Relative correction — a decaying offset is applied to the original motion, allowing the robot to temporarily shift its trajectory to complete the task. The offset gradually decreases at a decaying rate, ensuring the state remains within the target area temporarily, thereby reducing the need for frequent interventions. (b) Absolute correction — the operator fully overwrites the motion to provide high-quality demonstration data. However, this requires manually repositioning the robot to the end state after correction.}
     \label{fig:fig2}
     \vspace{-1em}
\end{figure}

\section{Methods}
In this section, we describe the methodology of the proposed the DRC method and robotic setup used to perform imitation learning and corrective feedback collection from expert.

\subsection{Robotic setup}
Our teleoperation system allows the operator to observe and intervene for seamless motion correction. The demonstrated system includes two 8-DoF cable-driven master manipulators with a foot pedal, as shown in Fig. \ref{fig:First_fig}. These manipulators are designed for high-DoF spatial dexterous manipulation. Additionally, three 8-DoF robotic arms, called CAFEs (Cable-driven Collaborative Floating End-Effectors) \cite{cheng_cafes} are developed and designed for agricultural applications. The CAFEs are mounted on a parallel rail system for horizontal movement, which is not used in this work for simplicity. Each CAFE is equipped with two RGB IMX 335 cameras, one on the gripper and one on top, for real-time vision feedback. Similar to \cite{chi2024universal}, two mirrors are placed beside the gripper to estimate depth information. Each CAFE computes its kinematics and communicates via the ROS2 network using an embedded Raspberry Pi and a servo controller.

The master controller provides position and orientation corrections (6-DoF), gripper angle control, and horizontal movement of the CAFEs (not used in this experiment). The foot pedal is used to trigger and switch between autonomous policy control and teleoperation mode. During the initial data collection stage, one of the teleoperation masters control a single CAFE using task / cartesian space mapping. In this stage, the foot pedal activates full teleoperation mode (manual control without policy involvement). Once the policy model is trained (behavior cloning) and deployed, the foot pedal is used to activate the policy and switch between teleoperated correction and autonomous policy control.

\begin{figure}[h]
    \centering
    \vspace{0.3cm}
    \includegraphics[width=1\columnwidth]{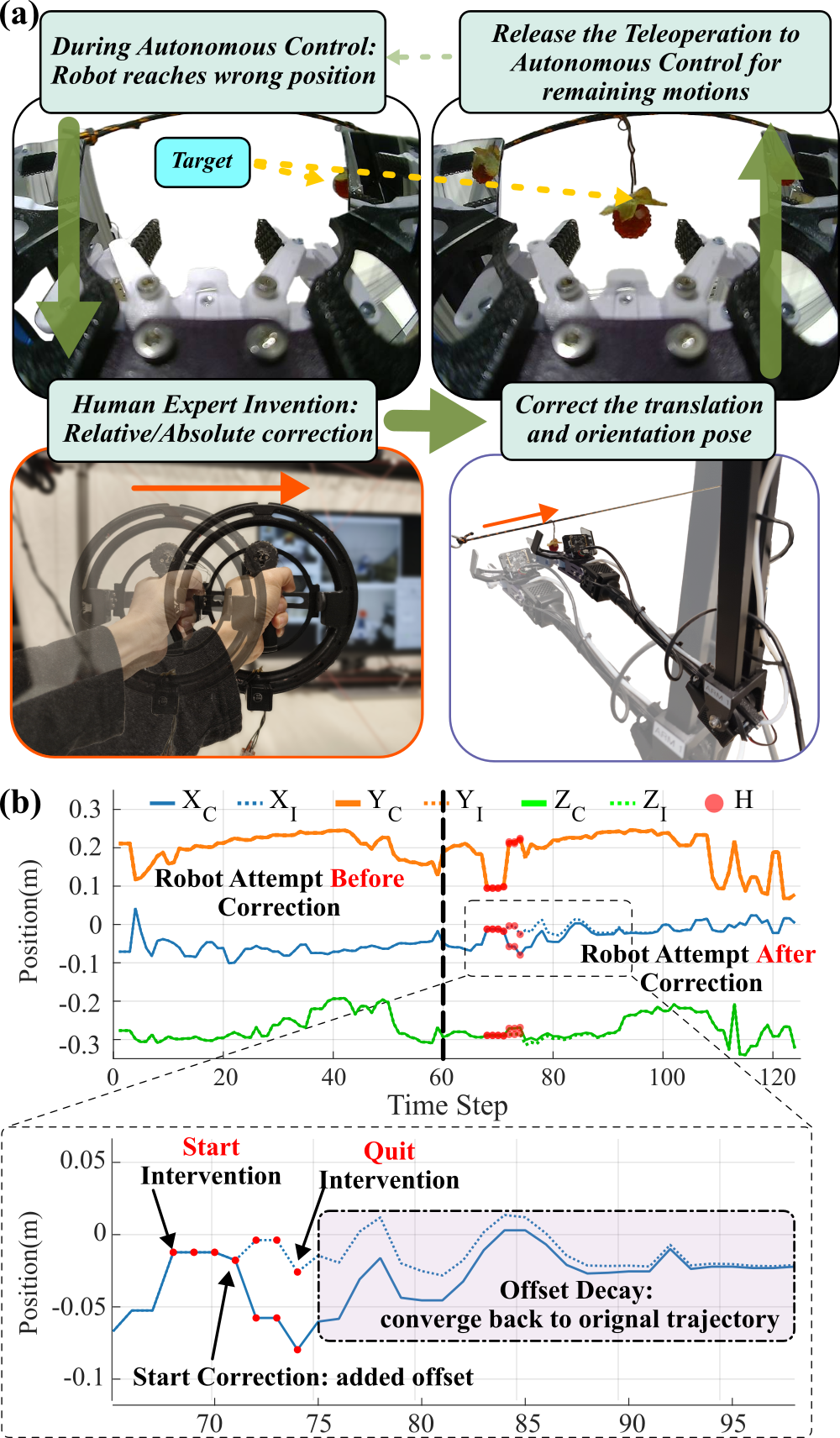}
    \caption{Human teleoperated correction when the policy model struggles. (a) Switching from autonomous policy control to human intervention to correct robot state via the cable-driven teleoperation handle. Once the correction is complete, control is returned to the robot. (b) Experimental Robot Trajectory Data – After a failed attempt, an expert corrects the robot’s motion using Direct Real-time Correction (DRC). The collected real-world data corresponds to Fig. \ref{fig:fig2} (a).}
    \label{fig:iros25_exp_setup}
    \vspace{-0.5cm}
\end{figure}

By using the foot pedal to switch operation modes, the operator can monitor the robot and determine when intervention is needed. As shown in Fig. \ref{fig:iros25_exp_setup}, the translation, orientation and the gripper of the robotic arm can be adjusted seamlessly during any time of the motion. A MATLAB-based user interface provides adjustable teleoperation parameters, including decay ratio, velocity, acceleration, and view angles. Due to MATLAB's interface limitations, the control frequency for computing the master's kinematics and capturing data is restricted to 12 Hz, while the robot policy model inference at 5 Hz.

\subsection{Decaying Relative Correction}


We introduce the proposed Decaying Relative Correction (DRC) method which enable experts to efficiently provide relative corrections to address incorrect robot behaviors. As illustrated in Fig. \ref{fig:fig2} (a), DRC applies corrective adjustments to robot actions for a limited duration during the early stages of correction, to drive the system state toward a desired target region. After the correction is removed by expert, the robot still remains within the target region for a certain number of subsequent steps due to the temporal persistence of the corrective offset. The correction then gradually diminishes following a decay process, allowing the robot to revert to its original trajectory.
The decay rate can be adjusted based on task requirements: a fast decay is suitable for dynamic, short-term tasks, while a slower decay is preferable for complex, long-duration tasks. The corrected action $a'_t$ under DRC is defined as:
\begin{equation}
a_t' =
\begin{cases}
a_t + v^{DRC},  \ \ \ \ \ \  \text{if expert inputs new} \ v^{DRC}  \\
a_t + v^{DRC} (1-r^{decay})^{t-t^{DRC}},   \text{otherwise}
\end{cases}
\label{eq:correction}
\end{equation}

where $a^{t}$ denotes the original action at time step $t$, $v^{DRC}$ is the corrective vector provided by the expert, $r^{decay}$ represents the decay rate, and $t^{DRC}$ is the time step at which the most recent $v^{DRC}$ was applied.  This formulation ensures the efficiency of correction in terms of the number of expert interventions required.

As shown in Fig. \ref{fig:fig2} (b), in contrast to DRC, an absolute correction strategy requires the expert to provide corrective inputs at every time step throughout the duration in which the robot deviates from the desired region and requires correction. This approach significantly increases the number of intervention steps needed from the expert, making it less efficient compared to the proposed DRC method.

\begin{figure*}[t]
    \centering
    \vspace{0.3cm}
    \includegraphics[width=0.9\linewidth]{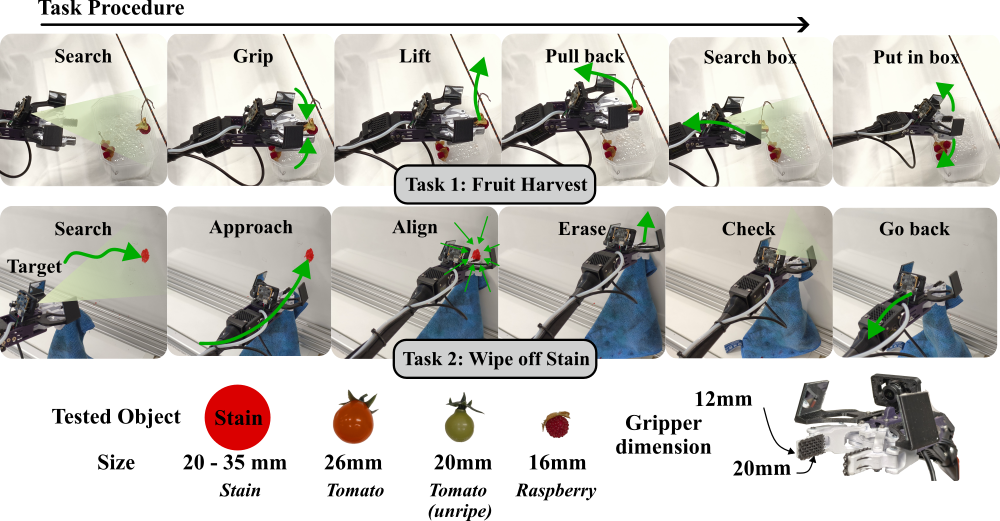}
    \caption{Experiments on Two Manipulation Tasks: harvesting and cleaning motion. The target object is randomly placed in a defined of area, and the robotic arm must locate it and execute the task autonomously. Small object sizes increase the task difficulty and highlights the effectiveness of the proposed method.}
    \label{fig:Task_step}
\end{figure*}

\subsection{Online Imitation Learning}


Based on corrections from expert, online imitation learning iteratively updates the robot control policy. When the robot executes incorrect actions, the expert acts as an online supervisor, correcting the robot actions. These corrected trajectories will be promptly utilized to update the robot control policy, leading to a progressively higher success rate and a corresponding reduction in errors. Particularly in the later stages, the recorded and corrected errors become increasingly valuable, enabling the model to leverage these high-quality data to achieve superior performance.

We present implementation details of our online imitation learning framework. A diffusion model based behavior cloning method \cite{chi2023diffusion} is applied in our framework. Initially, a dataset of demonstration $\mathcal{D}_P$ is collected using the master-slave teleoperation system (described in Fig. \ref{fig:First_fig}) to pretrain a policy model. Then the robot is deployed to perform tasks, with an expert providing corrections when the robot makes mistakes. Only the corrected trajectories are stored in a new dataset $\mathcal{D}_H$. After accumulating a sufficient number of corrected trajectories, the policy model is updated. 
To prevent the policy model from deviating excessively from previous model in each iteration and from overfitting to the limited amount of new dataset, both pretraining dataset $\mathcal{D}_P$ and new data with human intervention $\mathcal{D}_H$ are sampled in equal proportion for model updates. The policy model is updated by minimizing the following objective function:
\begin{align}
\theta &= \operatornamewithlimits{argmin}_{\theta}  \mathbb{E}_{(s,a) \sim \mathcal{D}_P}  \left\| \pi_\theta(s) - a \right\|^2 \notag \\
&\quad + \mathbb{E}_{(s,a) \sim \mathcal{D}_H} \left\| \pi_\theta(s) - a \right\|^2,
\label{eq:policy_update}
\end{align}

where $\pi_\theta$ denotes the policy model with parameters $\theta$, and $s$ and $a$ represent the robot state and action, respectively. $s$ consists of two frames of images from the robot’s top and wrist cameras, concatenated with the gripper’s 6-DOF position and state. The predicted $a$ is a sequence of 16 steps, each a 7-DOF vector representing the gripper’s 6-DOF position and state. Every 8 steps are executed before the diffusion model performs the next inference.

The training of diffusion policy model is based on the collected grasping demonstration. Images from two cameras are resized into 320x240 as inputs of the model, during training, images are randomly cropped images into 288x216 as data augmentation. The training parameter settings are same as in the CNN-based diffusion policy in \cite{chi2023diffusion} . The first model is pretrained for 1500 epochs. For subsequent online updates, the model is trained for 500 additional epochs, with training occurring every time 10 new corrective trajectories are collected.





\section{Experimental Setup}
\label{sec:exp_set}

In this section, we introduce the experiments about two manipulation tasks to evaluate the efficiency of proposed decaying relative correction and performance of whole online imitation learning framework, along with the corresponding metrics. Since both tasks are designed for single robot arm, all data collection including demonstration and correction is deployed using a single robot arm with a single cable teleoperation handle.


\subsection{Manipulation tasks}

\subsubsection{Task 1 - Harvesting of Hook-Suspended Artificial Raspberries}

As shown in the upper sub-figures of Fig. \ref{fig:Task_step}, we investigate robotic harvesting of an artificial raspberry suspended from a rope using a hook. Unlike real raspberry harvesting, which typically requires only a downward pulling motion, this scenario introduces additional manipulation challenges. The gripper must first align with and grasp the raspberry, then lift it slightly to disengage the hook from the rope before completing the harvest. Then the robot moves horizontally backward to clear the hanging point before moving to the collection box, where the raspberry is placed. The target fruit is randomly suspended within a certain area on the rope for data collection and testing, while the placement box remains fixed. Note that this test presents a challenging fine manipulation task, as the gripper measures only 12 × 20 mm, while the target raspberry is a 16 mm-diameter sphere. A misalignment of 10 mm, which can result from factors such as the robot's backlash or the deformation of the suspended rail, significantly increases the likelihood of grasping failure. To evaluate success beyond a binary metric, task performance is scored on a three-level scale: 1 for full task success, 0.5 for successful object pickup only, and 0 for failure.

To evaluate adaptability to unseen objects within the online imitation learning framework, we introduce two additional artificial fruits in the harvesting task: a green (unripe) cherry tomato and an orange cherry tomato. For both new objects, we conduct experiments using the well-trained policy for raspberries, as well as an updated policy that incorporates expert correction data specific to the new object.



\subsubsection{Task 2 - Stain Removal on a Whiteboard}
The second experiment is illustrated in lower sub-figures of Fig. \ref{fig:Task_step}. In this task, the robot, equipped with a wiping cloth in its gripper, must search for and approach a red stain on a whiteboard. The stain is approximately circular, with a randomly drawn diameter ranging from 20 to 35 mm. The robot will first find and then align the front tip, covered by the cloth, with the stain and moves forward until it makes contact with the whiteboard. The depth or contact information is estimated by two mirrors mounted besides the gripper. The robot then moves upward to erase the stain. Finally, the robot checks if any stain remains and repeats the erasure process if necessary.
The target stain is randomly drawn within a specified area on the whiteboard for data collection and testing.  The task is scored using one of three possible values: 1 for complete erasure of the stain, 0.5 for erasing more than 50\% of the stained area, and 0 for all other cases. 


\subsection{Implementation and Evaluation Metrics}
\label{sec:expset_2}

To evaluate the efficiency of the propose DRC method, we compare it with a standard absolute correction method based on the expert intervention rate (percentage), defined as $n_\text{intervention} / n_\text{traj} $. Here, $n_\text{intervention}$ represents the number of steps in which the human expert provides corrective signals, while $n_\text{traj}$ denotes the total number of steps in the entire trajectory. This intervention rate is computed across 10 corrected trajectories in each iteration of online imitation learning.

To assess the effectiveness of the online learning framework in improving the policy model, the success rate is used as a performance metric. Online imitation learning is performed for 3 rounds, with each round incorporating 10 corrections for using DRC and these trajectories are used to update the policy model. The pretrained model and the three updated models are then evaluated through 20 repeated trials for Task 1 and 10 repeated trials for Task 2 to compute the success rate.

\section{Experimental Results}

In this Section we first compare the effects of the two relative correction methods before deploying the optimal one for online training with expert correction for the two tasks specified in Section  III. 

\subsection{Efficiency of Decaying Relative Correction}

\begin{figure}[t]
     \centering
     \vspace{0.3cm}
\includegraphics[width=0.85\columnwidth]{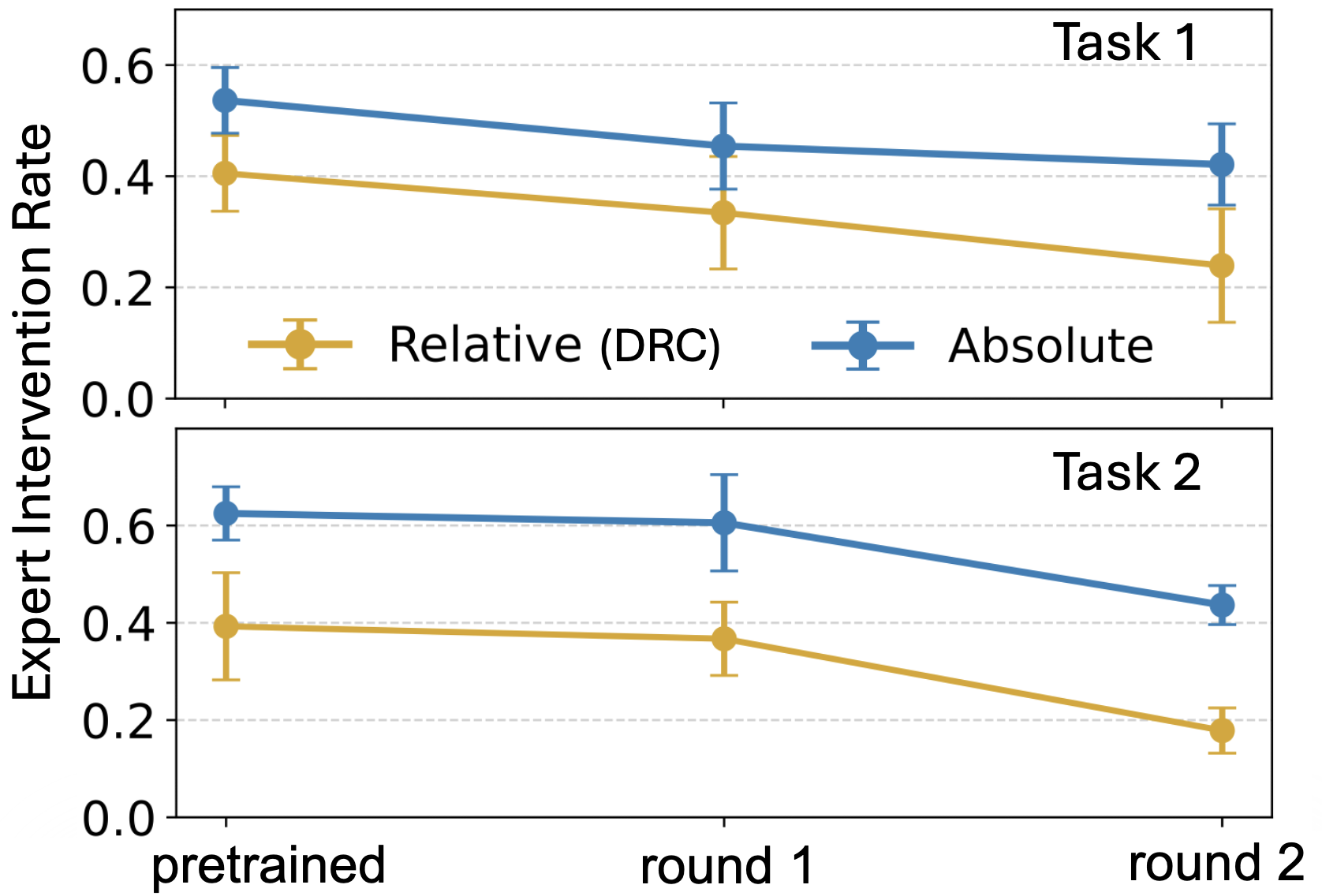}
     \caption{The expert intervention rate (ratio of time steps during the task where the expert is providing corrections) for the two tasks for the pre-trained policy model (trained on 20 demonstrations), and 2 further rounds of online imitation learning, where 10 additional corrected trajectories are used to update the policy model.}
     \label{fig:control_rate}
     \vspace{0em}
\end{figure}

We use intervention rate (see Sec. \ref{sec:expset_2}) to evaluate the efficiency of our proposed DRC method. We measure this for 10 different repeats of the two tasks for three different rounds of the policy model training: a pre-trained policy, and the policy model in round 1 and 2 are updated each with an additional 10 corrections, and thus we expect an improved performance of the policy.  In Fig.~\ref{fig:control_rate} we show the expert intervention rate for the two tasks for these three rounds of performance, with the mean and standard deviation given for 10 repeats.  For both tasks we see that DRC method consistently requires less intervention that the absolute one, typically around 30\% less. With each additional round of training the expert intervention rate decreases for both methods as the policy model improves and less intervention steps is required. Within this the ratio between the rates of the two different methods stays approximately constant.  This reflects the longer time the expert must be involved for each absolute intervention to continuously correct the robot until it becomes effective again, where as for the RDC approach, after the intervention is made, and the time delay used to  temporarily keep the robot within the target working area. Given the increased efficiency for expert of the relative intervention method this is used for the remainder of the experiments. 


\begin{figure}[tb!]
     \centering
     \includegraphics[width=0.85\columnwidth]{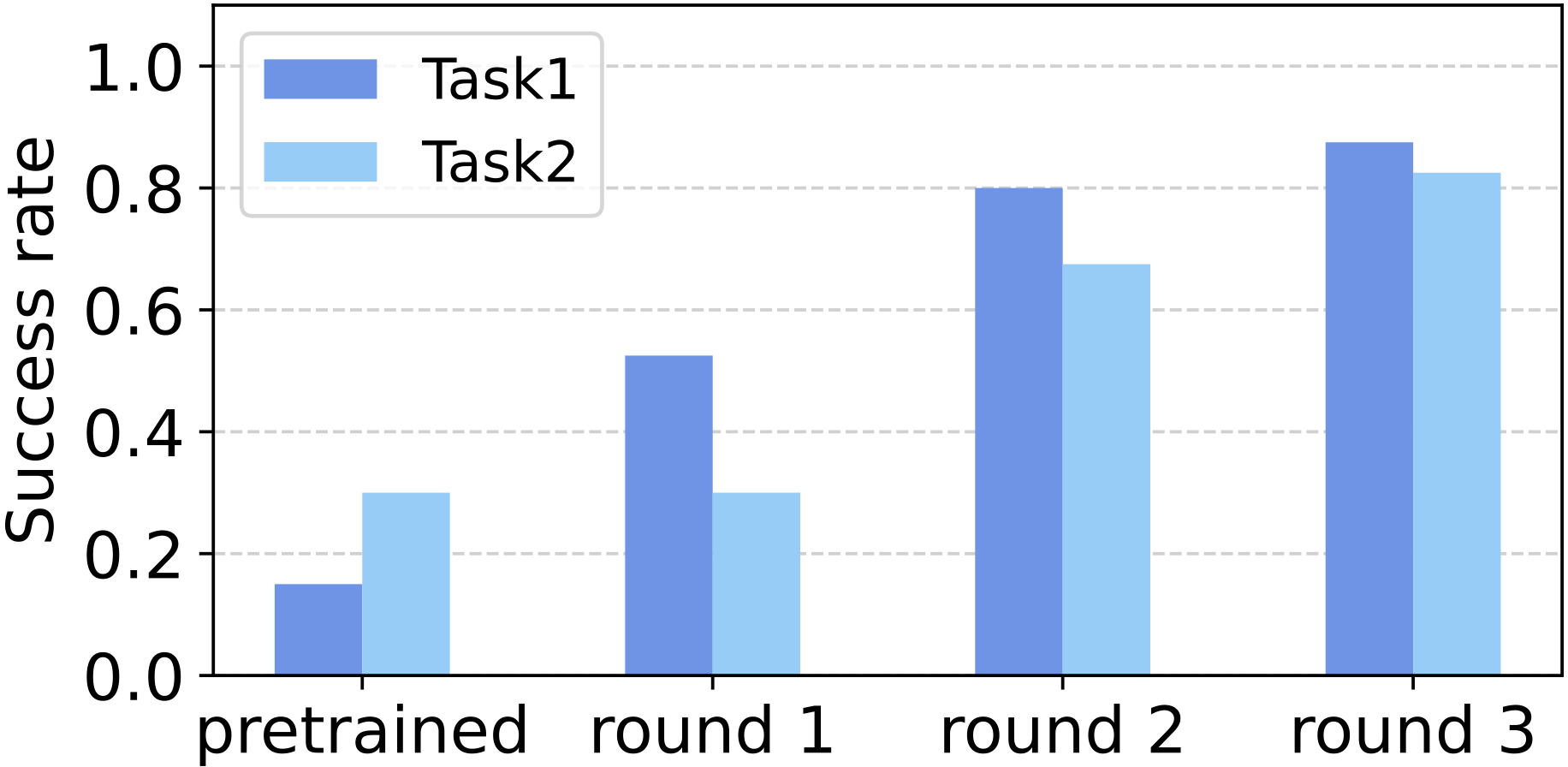}
     \caption{The measured success rate for Task 1  and Task 2 for the pre-trained policy model and  3 further rounds of online imitation learning.}
     \label{fig:iter_improve}
     \vspace{-1em}
\end{figure}

\subsection{Online Imitation Learning via Expert Supervision}

For the two tasks, we use the most efficient intervention approach, namely DRC, for expert supervision and correction to perform online imitation learning. For each task we start with a pre-trained policy which is trained on 20 expert demonstrations of the task. Following this we have three rounds of gathering an additional 10 demonstrations of the task with expert corrections. After each round, the policy is re-trained, deployed and the success rate measured.  The success rate is calculated based on success defined as in Sec. \ref{sec:exp_set} within the time period of one minute.

The results, illustrated in Fig.~\ref{fig:iter_improve}, show that for Task 1 with each round of online imitation learning the success rate improves, from the initial pretraining success rate of 0.15, through to  close to a 0.9 success rate by round 3.  For Task 2 in round 1 of policy updating, there is initially minimal improvement, we hypothesize this is due to the low-quality demonstration data such that the stain is not well recognized in the camera. However, round 2 and 3 show considerable improvement after we make sure the stain is visible in the both mirrors (for better depth estimation) on the gripper during new demonstrations, to achieve a similar success rate of above 0.8 by round 3.  Thus, the use of expert user feedback when the policy model makes mistakes leads to an increase in success rate with a limited number of demonstration trajectories. Unlike training from scratch with a similar number of full trajectories, this approach requires only minimal expert intervention steps, as shown with the intervention rate given in Fig.~\ref{fig:control_rate}.


\subsection{Task Transfer via Online Imitation Learning}

\begin{figure}[t]
     \centering
     \vspace{0.3cm}
     \includegraphics[width=0.85\columnwidth]{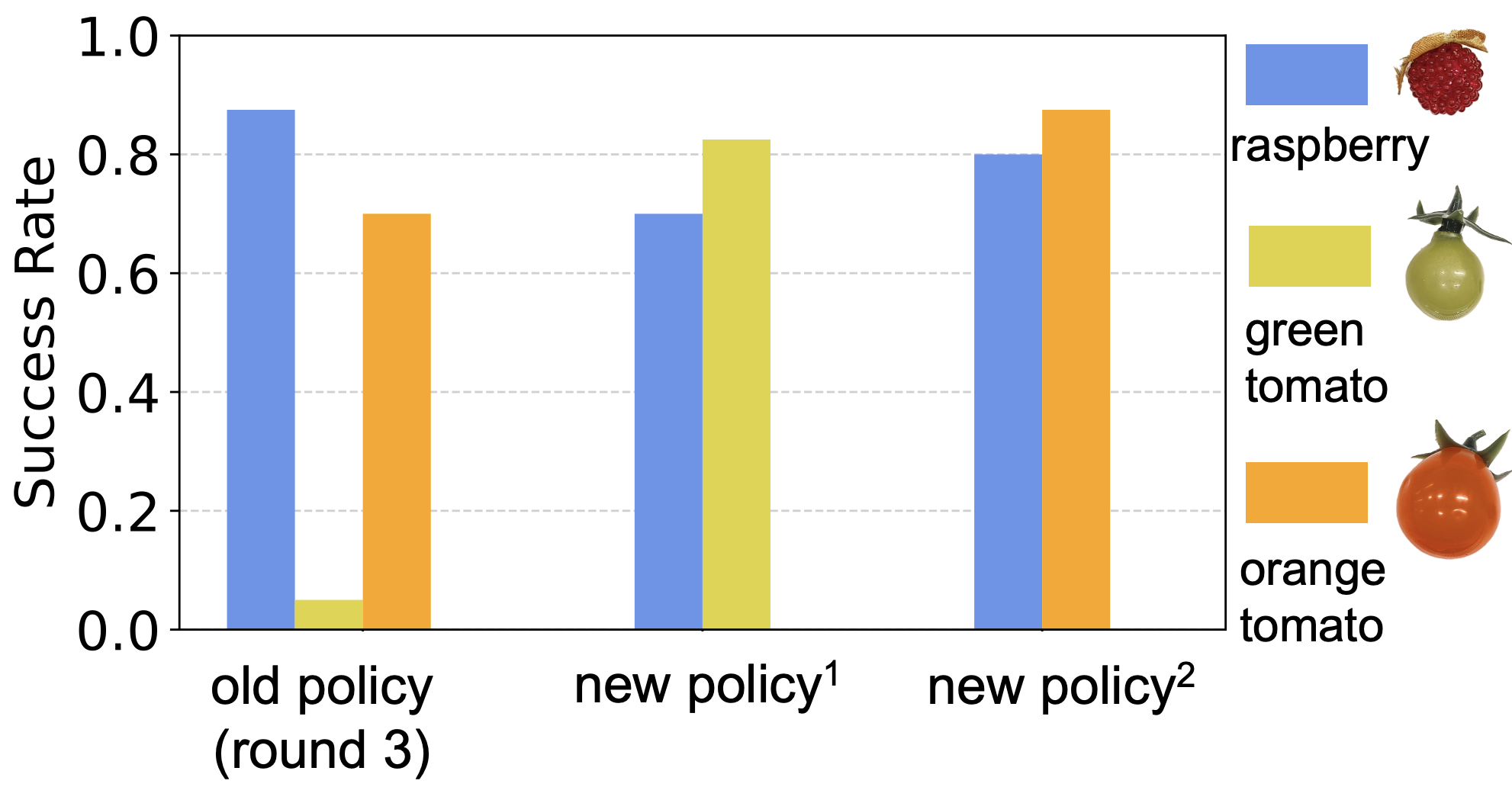}
     \caption{The success rate of the tasks is report for three objects (raspberry used in Task 1, green tomato and orange), first for the policy used in Task 1, round 3, and then a new policy is generated using 10 trajectories with human correct to create a new policy for task transfer for both the green tomato (new policy 1) and the orange tomato (new policy 2). }
     \label{fig:task1_transfer}
     \vspace{-1em}
\end{figure}

Another use of the human correction and online imitation learning is to utilize the feedback to adapt the task, to enable transfer to a similar task. There must be some similarity between the tasks such that this is more effective than training from scratch. We consider the task transfer from Task 1 (raspberry harvesting) to harvesting objects of different sizes and colors.  First to an orange cherry tomato which has more visual similarity to the raspberry, and secondly, a green (unripe) cherry tomato which has less similarity in color.  

\begin{figure}[t]
    \centering
    \vspace{0.5cm}
    \includegraphics[width=1\columnwidth]{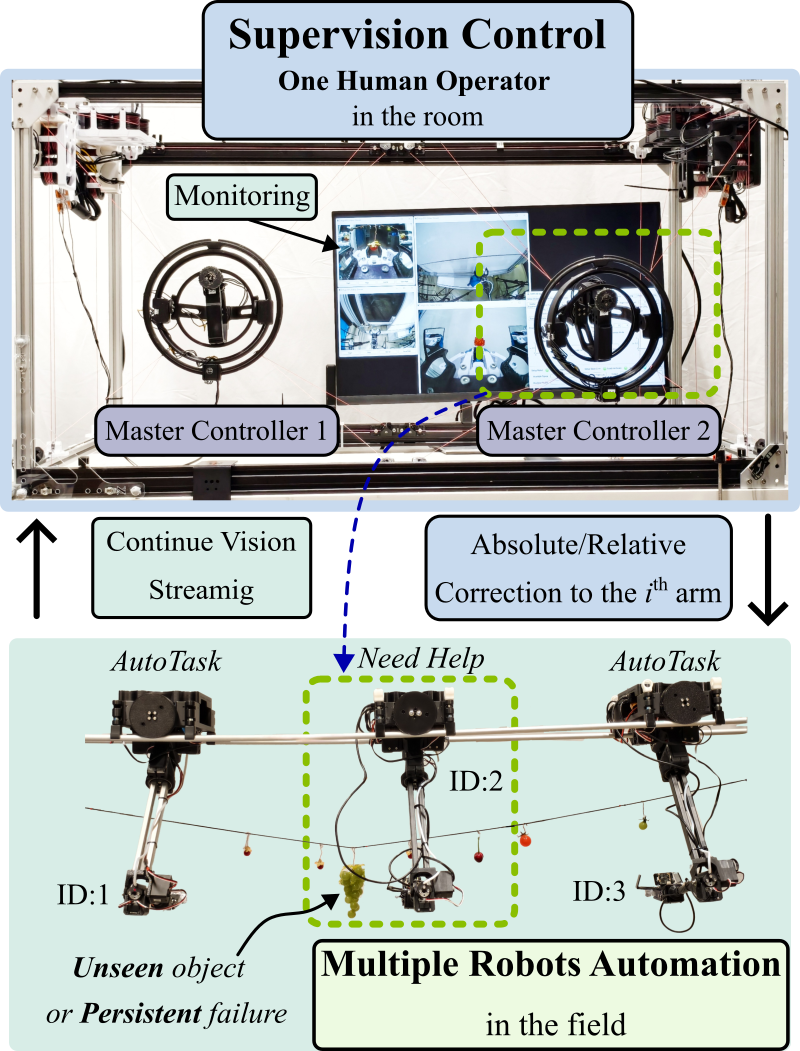}
    \caption{Expansion of the teleoperated correction system to enable a single expert to supervse multiple arms and provide corrections. This facilitates the sharing of corrective trajectories, which are used to update the policy model. The improved control policy is then applied across all arms, enhancing overall performance.}
    \vspace{-0.6cm}
    \label{fig:teleop_sch}
\end{figure}

We initially use the Task 1 control policy from round 3 to attempt to harvest all three object types.  As shown in Fig.~\ref{fig:task1_transfer} the performance of this policy is highest for the raspberry, with a success rate of approximately 0.9.  For the two out of distribution objects, the more similar orange tomato has a higher success rate of 0.7, whereas the green tomato has a low success rate of less than 0.1.  Using online imitation learning, 10 new trajectories with corrections are collected for each of the two objects, followed by training for two new policies.  For the updated new policy 1, retrained on green tomato data, with only this small amount of additional data we see an increase for 0.1, to a success rate greater than 0.8 for the green tomato, however, the performance for the raspberry drops marginally.  Similarly for the new policy for the orange tomato there is an increase in the performance for the tomato whilst the raspberry performance has a slight decrease.  This results demonstrate how human correction and online retraining can enable the policy model to updated to work for multiple objects. This can be valuable to efficiently generalize a control policy, or adapt the policy with a change in the task. 

\vspace{-0.1cm}

\subsection{Expansion to Multi-arm Correction}

The use of DRC reduces the amount of time taken by the expert to provide the correction signal, potentially enabling a single expert to supervise and correct multiple robots via a single teleoperation system. In Fig.~\ref{fig:teleop_sch} we demonstrate how the robot system (based upon the CAFEs~\cite{cheng_cafes}) can be extended to incorporate multiple arms. In this setup, corrections can be applied selectively to individual robots, while the newly recorded trajectories are subsequently used to retrain the control policy, which can then be deployed across all robots. In other words, errors and learned experiences from individual robots can be shared to enhance the overall performance of all robots.

\section{Discussion \& Conclusion}

Enabled by a teleoperation system that allows rapid switching between an autonomous controller and user input, we compare two methods for expert correction of control policies and their effectiveness in online retraining for rapid teaching and fine-tuning of tasks. Specifically, we compare Decaying Relative Correction (DRC) with an absolute correction method and demonstrate that the proposed DRC approach reduces the required human intervention rate by approximately 30

Enabled by a teleoperation system that allows rapid switching between an autonomous controller and user input, we compare two methods for expert correction of control policies and how this can be used in an online learning framework for rapid teaching and fine-tuning of tasks.  Specifically, we compare a relative (DRC) and absolute correction method, and demonstrate how the proposed DRC correction requires approximately 30\% less intervention rate from a human expert to correct the trajectory. Using this relative correction method, we show that collecting expert corrections for two manipulation tasks leads to a rapid increase in the success rate of the control policy when combined with online imitation learning. Furthermore, for similar tasks, this approach can also enable task adaption or generalization through an expert user providing corrections to adapt a controller for a task with unseen objects.

Further work is now required to further optimize the DRC correction method, such as incorporating a self-adaptive decay rate, and to evaluate the approach across a wider range of manipulation tasks. Expanding on our initial demonstration of multi-arm correction (Fig.~\ref{fig:teleop_sch}), the potential to leverage the efficiency of our method of corrections to enable correction of multiple robotic arms should be investigated, allowing errors and learned experiences from individual robots to be shared and improving the overall performance of the system.




\section*{Acknowledgements}
This research project was supported by the European Union’s Horizon 2020 research and innovation programme under the Marie Skłodowska-Curie grant agreement No. 945363.

\begingroup
\LARGE  
\bibliographystyle{IEEEtran}
\bibliography{refs}
\endgroup

\end{document}